\documentclass[a4paper,fleqn]{cas-dc}
\usepackage{graphicx}
\usepackage[numbers]{natbib}
\usepackage[ruled,linesnumbered]{algorithm2e}
\usepackage{amsmath}
\usepackage{autobreak}
\usepackage{cleveref}
\usepackage{stfloats}
\usepackage{float}
\usepackage{enumitem}
\usepackage{caption}
\usepackage{subcaption}
\DeclareCaptionSubType*[Alph]{table}
\DeclareCaptionLabelFormat{mystyle}{Table~\bothIfFirst{#1}{ }#2:~}
\crefname{section}{§}{§§}
\def\tsc#1{\csdef{#1}{\textsc{\lowercase{#1}}\xspace}}
\tsc{WGM}
\tsc{QE}
\tsc{EP}
\tsc{PMS}
\tsc{BEC}
\tsc{DE}

\begin{document}
\let\WriteBookmarks\relax
\def\floatpagepagefraction{1}
\def\textpagefraction{.001}
\shorttitle{Knowledge-Based Systems Submission}
\shortauthors{WANG et~al.}

\title [mode = title]{Empathetic Response Generation through Graph-based Multi-hop Reasoning on Emotional Causality} 

\author{Jiashuo WANG}[orcid=0000-0002-8254-8138]
\ead{jessiejs.wang@connect.polyu.hk}

\author{Wenjie LI}[orcid=0000-0002-7360-8864]
\ead{cswjli@comp.polyu.edu.hk}

\author{Peiqin Lin}[orcid=0000-0003-2818-3008]
\ead{lpq29743@gmail.com}

\author{Feiteng Mu}[orcid=0000-0003-1337-0388]
\ead{feitengmu.mu@connect.polyu.hk}

\address{Department of Computing, The Hong Kong Polytechnic University, Hong Kong}

\begin{abstract}
Empathetic response generation aims to comprehend the user emotion and then respond to it appropriately. Most existing works merely focus on what the emotion is and ignore how the emotion is evoked, thus weakening the capacity of the model to understand the emotional experience of the user for generating empathetic responses. To tackle this problem, we consider the emotional causality, namely, what feelings the user expresses (i.e., emotion) and why the user has such feelings (i.e., cause). Then, we propose a novel graph-based model with multi-hop reasoning to model the emotional causality of the empathetic conversation. Finally, we demonstrate the effectiveness of our model on \textsc{EmpatheticDialogues} in comparison with several competitive models.

\end{abstract}


\maketitle

\section{Introduction}
Empathetic responding, which aims to understand the emotional experience of the user and then reply to acknowledge it appropriately, has benefited a wide range of downstream applications, such as medical dialogue systems \cite{eide2004listening}, counseling conversation \cite{perez2017understanding} and social chatbots \cite{zhou2020design}. Recent years have witnessed the emergence of empathetic dialogue systems, and some methods \cite{moel,majumder2020mime,li2019empgan} have been proposed for the task of empathetic response generation and achieved promising results. Specifically, to better understand the emotional experience of the user, MoEL and MIME \cite{moel,majumder2020mime} utilize the emotional labels, and EmpDG \cite{li2019empgan} uses both coarse-grained labels annotated by human and fine-grained emotional words identified by the sentiment lexicons. However, these methods only focus on what the emotion is when encoding the conversation history, which is insufficient to fully understand the emotional experience of the user.

From the aspect of psychology, emotional cause is recognized as an important feature in the analysis of human emotion \cite{wierzbicka1999emotions}, and itself should be an integral component of emotional experience \cite{whatisemo}. Therefore, to understand the emotional experience for generating empathetic responses, it is essential to figure out the forming process (i.e., cause) of the present feelings (i.e, emotion).

To this end, we propose a novel model, named \textbf{G}raph-based multi-hop \textbf{R}easoning on \textbf{E}motional \textbf{C}ausality (\textbf{\textit{GREC}} for short), which introduces emotional causality analysis to help understand the emergence of the user's emotion and facilitate empathetic response generation with such knowledge. Specifically, to represent the user's emotional experience embedded in the conversational history, we first construct a series of emotional causality graphs through multi-hop reasoning over the external commonsense graph. Then, we adopt the multi-layer Graph Convolutional Networks (GCN) \cite{gcn} to model the emotional causality graphs. The encoded graphs are used to serve for the response generation in two ways. On the one hand, the fused graph-aware representation is fed into the decoder to generate a generic vocabulary distribution for the empathetic response. On the other hand, to explicitly exploit the information of emotional causality, multi-hop reasoning on the emotional causality graphs is adopted to obtain the vocabulary distribution over the nodes of the graphs at each step of generation, which affects the final vocabulary distribution.

The main contribution of this paper can be summarized as follows:

\begin{enumerate}[label=(\arabic*)]
\item It integrates emotional causality reasoning into empathetic response generation, providing an inner and deeper understanding of the user's emotional experience. To the best of our knowledge, this is the first work on the topic of empathetic conversation.
\item It develops a novel multi-hop graph-based reasoning and generation approach to capture the emotional causal relations among utterances and model the process of emotion forming. Reasoning on graphs improves the explainability of the model to some extent.
\item It verifies the effectiveness of the proposed model as well as the contribution of each component of the model. The experimental results demonstrate that \textit{GREC} with emotional causality reasoning enhances the empathetic expression of replies compared to baselines. 
\end{enumerate}  

\section{Related Work}
\subsection{Empathetic Response Generation}
Recent years have witnessed the emergence of empathetic dialogue systems \cite{ma2020survey}. Similar to the task of emotional response generation, empathetic response generation also focuses on the emotional expression of users. However, there is a remarkable difference between these two tasks. Emotional response generation aims to generate responses with specific emotions \cite{shi2018sentiment, zhong2019affect, ecm, song2019generating,mojitalk}, while empathetic response generation emphasizes response generation by understanding the emotional experience of the user. Several works have already attempted to make chatbots more empathetic. Specifically, Rashkin et al. have proposed a new dataset and benchmarked several existing models on it \cite{empatheticdialogues}. With a transformer-based encoder-decoder architecture, MoEL \cite{moel} softly combines different outputs from several emotional decoders and generates the final response. MIME \cite{majumder2020mime} assumes that empathetic responses often mimic the emotions of the speakers to some degree, and its consideration of polarity-based emotion clusters and emotional mimicry shows to improve empathy and contextual relevance of responses. EmpDG \cite{li2019empgan} exploits both dialogue-level and token-level emotion to capture the nuances of the speaker's emotion. Moreover, it also involves the speaker's feedback to achieve better expression of empathy with an interactive adversarial learning framework. Based on GPT, CAiRE fine-tunes on empathetic dialogue corpus after pretraining a language model with large corpus and multiple objectives \cite{lin2019caire}. Besides, it is also a promising research direction to involve empathy in various dialogue systems to increase the user experience. For instance, audio-based dialogue systems consider voice tone, intonation, and content of speakers to generate empathetic responses \cite{eide2004listening,torres2019spoken}. Fung et al. investigated empathetic dialogue for booking systems with a latent variable to store the dialog state \cite{fung2018empathetic}. In this paper, we focus on empathetic response generation for text-based chit chat. 

\subsection{Emotion Cause Analysis}
With deeper exploration of human emotions and feelings, more attention is turned to inner emotional information (e.g., experiences, causes, and consequences of an emotion) \cite{alm2005emotions} from apparent emotional information (e.g., emotion categories) \cite{aman2007identifying,bakshi2016opinion}. The stimuli and cause of emotion have been explored in recent ten years in the field of natural language processing. Emotion cause analysis was first proposed as a new task \cite{lee2010text} along with a linguistic rule-based method and a manually annotated small scale dataset \cite{lee2010emotion}. After several works \cite{gui-etal-2016-event,xia2019emotion} refines the concept of emotion cause, neural based models were come up with. Gui et al. \cite{gui2017question} extracted the emotion cause by considering emotion cause extraction as a reading comprehension task in question-answering. Xia et al. \cite{xia2019emotion} proposed a two-step approach, which first performs individual emotion extraction and cause extraction via multi-task learning, and then conducts emotion-cause pairing and filtering. A RNN-transformer based hierarchical network, named RTHN, was also proposed to extract emotion cause at the level of clause \cite{rthn}. However, the emotion-cause relation has not been addressed in the existing methods for empathetic response generation. In this paper, we adopt the aforementioned RTHN \cite{rthn} to discover the emotion cause relations from conversations and leverage this inner emotional evidence to generate empathetic responses. 

\subsection{Dialogue Systems with Multi-hop Reasoning over Commonsense Knowledge Graphs}
Multi-hop reasoning over external knowledge graphs has been proved an effective approach to involve rich semantic information and enable the explainability of models. On the one hand, informative commonsense observations and relations in the structured knowledge boost studies on knowledge fusion between semantic and symbolic space for reasoning \cite{ji2020survey,zhou2018commonsense}. On the other hand, paths and flows provide evidences for knowledge selection and deduction, making systems explainable. Therefore, multi-hop reasoning is popular in various downstream tasks. Specifically, KagNet \cite{lin2019kagnet}, a model based on graph convolutional networks and LSTMs with a hierarchical path-based attention mechanism, is proposed for the task of commonsense question answering. For reading comprehension based QA, a multi-hop pointer-generator model was proposed, along with a selection algorithm to extract knowledge from ConceptNet and an attention mechanism to generate answer \cite{bauer2018commonsense}. For conversation generation, Liu et al. presented an augmented knowledge graph based open-domain chatting machine \cite{liu-etal-2019-knowledge}. Moreover, multi-hop reasoning has also been used in long text generation, such as story completion, review generation, and description generation \cite{ji2020language,zhao2020graph}. Our work preforms multi-hop reasoning over external commonsense knowledge to imitate the processes of emotion formation for better understanding human emotion.

\section{Methodology}
\begin{figure*}
    \centering
    \includegraphics[width=\textwidth]{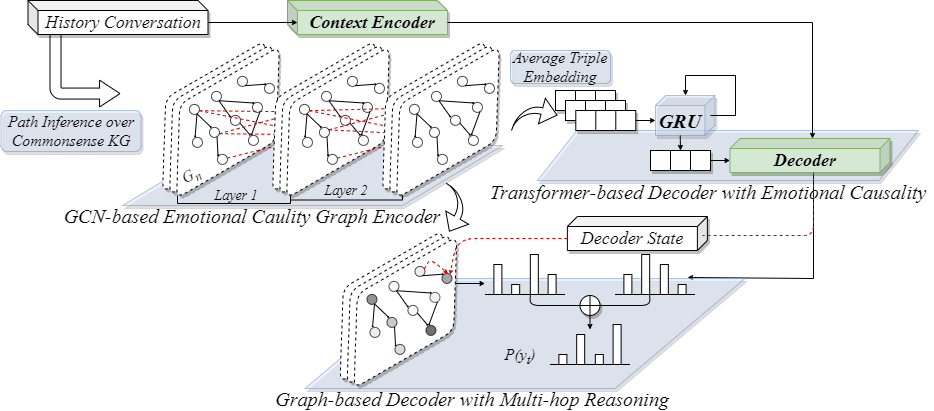}
    \caption{This is an overview architecture of \textit{GREC}. Apart from the transformer context encoder to encode the conversation context, a GCN encoder encodes the emotional causality graphs after the construction through multi-hop reasoning over the commonsense knowledge graph. At the decoder side,  a transformer-based decoder augmented with emotional causality and a graph-based decoder with multi-hop reasoning work together to the response generation.}
    \label{overview}
\end{figure*}

The overview architecture of our proposed model is shown in \figureautorefname~\ref{overview}. Based on a typical transformer-based Seq2Seq model, we use the emotional causality to augment both the encoder and the decoder with a graph-based approach.

For the encoder, we construct a series of emotional causality graphs for the conversation through multi-hop reasoning over ConceptNet (\cref{sec:construct}), and then apply a multi-layer GCN encoder to model these emotional causality graphs (\cref{sec:graph-encoder}). Besides, we use a typical transformer encoder for context understanding (\cref{sec:trs-encoder}).

For the decoder, we combine both the sequential information obtained by the transformed-based conversation context encoder and the structural knowledge obtained by the graph-based emotional causality encoder to computes the generic vocabulary distribution (\cref{sec:trs-decoder}). Additionally, a graph-based decoder with multi-hop reasoning computes the distribution over the concepts of the graph to enhance the generic vocabulary distribution for better generating the response (\cref{sec:graph-decoder}). 
\subsection{Emotional Causality Augmented Encoder}\label{sec:encoder}
\subsubsection{Emotional Causality Graph Construction} \label{sec:construct}
To understand the emotional causality of the user's emotional experience in the conversation context, we construct a series of emotional causality graphs based on the given context. Specifically, we first identify the emotion concepts and the cause concepts in the dialogue context, and then reason out the relations between emotion concepts and cause concepts from the external commonsense knowledge graph, namely, ConceptNet, through a multi-hop strategy for constructing the emotional causality graphs.

To obtain the emotion concepts and cause concepts, we need to first identify the emotion clauses and cause clauses of the conversation context. For the emotion clauses, we directly regard the clauses of the immediate utterance $\mathcal{Q}$ as the emotion clauses, since we aim to generate the specific response with empathy for the immediate utterance. Here, we construct multiple emotional causality graphs for different clauses of the immediate utterance rather than a single graph for the immediate utterance because there may be different causes of the immediate utterance's emotional clauses. Then, for each emotion clause, we extract its cause clauses from the conversation context by applying an emotion cause extraction model, namely, RTHN \cite{rthn}, which can identify whether a clause is the emotion cause of another clause in a given context. The cause clauses are simply flatted as the cause context of each emotion clause. Given the emotion clauses, we select verbs, nouns, adjectives, and adverbs in both emotion clauses and cause clauses as the emotion concepts and cause concepts, respectively. 

\begin{figure}[ht]
	\centering
	\includegraphics[width=0.7\linewidth]{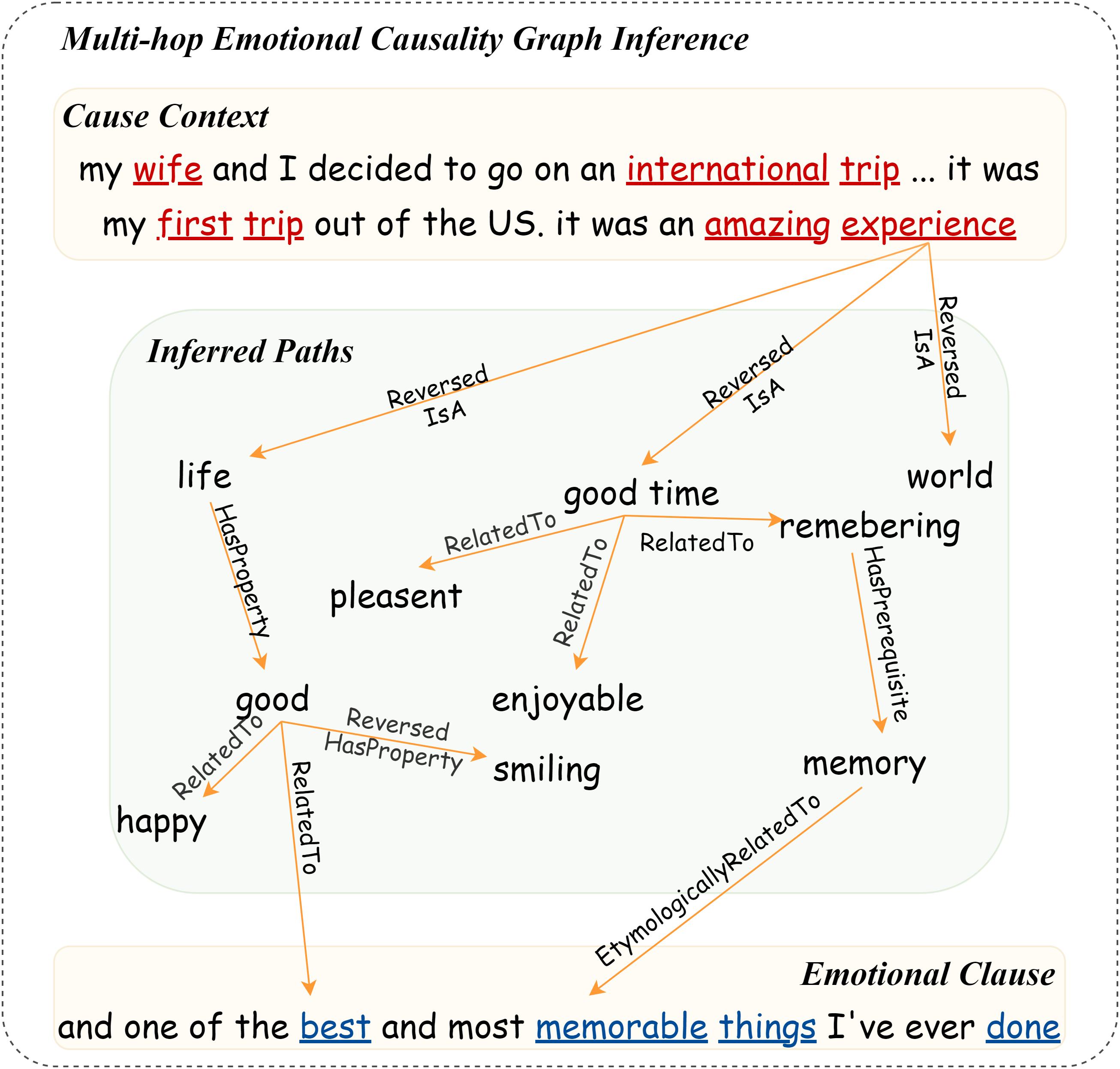}
 	\caption{This is an example that illustrates the process to deduce causal paths over an external commonsense graph starting from the cause concept \textit{experience}. \textsc{\textcolor{red}{\textsf{\textit{\underline{Underlined terms in red}}}}} correspond to the cause concepts in the cause context, while \textsc{\textcolor{CornflowerBlue}{\textsf{\textit{\underline{underlined terms in blue}}}}} correspond to the emotion concepts in the emotional clause. The \textsc{\textcolor{OliveGreen}{\textit{green space}}} represents a segment of the external commonsense knowledge graph. All the available concepts during this process are maintained to involve the rich semantic information. We preform multi-hop emotional causality inference on clause pair with a cause-emotion relation via a series of similar deduction processes from all cause concepts.}
	\label{fig:ECG}
\end{figure}

To construct the emotional causality graphs based on the emotion concepts and cause concepts, we adopt multi-hop reasoning over ConceptNet to infer semantic paths from emotion concepts to cause concepts, as shown in \figureautorefname~\ref{fig:ECG}. The detailed process of deriving intermediate concepts and relations which have the potential to help connect cause concepts with emotion concepts is shown in \algorithmcfname~\ref{alg:Framwork}. Specifically, we use a concept stack \textbf{$S$} to collect all candidate concepts to be in the emotional causality graph. The concept stack \textbf{$S$} is initialized with all cause concepts in $C_{c a s}^{(i)}$. At each hop $h$, we pop concepts from \textbf{$S$}, and collect their neighbour concepts that have not been passed (line 6-10). Then, we select the top $K$ concepts from all neighbours concepts according to semantic similarity and push them into the concept stack \textbf{$S$} (line 11-17). Here, we regard the cosine similarity between the word embeddings as the semantic similarity of the two concepts. If there is an emotion concept $v_{e}$ in the concept stack \textbf{$S$}, we pop $v_{e}$ directly without collecting neighbouring concepts of $v_{e}$. After $H$ hops, we obtain the emotional causality graph. The intermediate concepts and edges are kept in the emotional causality graph to simulate the behaviour that humans think of various terms in their brains to construct a sentence when chatting.

\begin{algorithm}[t]  
  \caption{Framework of Emotional Causality Graph Construction by Multi-hop Reasoning over \textit{ConceptNet}}
  \label{alg:Framwork}  
    \KwIn  {\textit{ConceptNet}, $G_{C}$; \\
    The sets of cause concepts and emotion concepts: $V_{cas}=\left\{{v_{1}}^{cas}, \ldots, {v_{M}}^{cas}\right\}$, $V_{emo}=\left\{{v_{1}}^{emo}, \ldots, {v_{N}}^{emo}\right\}$.}
    \KwOut {An emotional causality graph, $G$.}
    \BlankLine
    Initialize an emotional causality graph $G=\left(v_{cas},\emptyset\right)$;\\
    Initialize a concept stack $S$ with cause concepts;\\
    Initialize an empty temporal triple list $T$;\\
    \For{h from 1 to H}{
        \While{$S$ is not empty}{
            pop $v_{i}$ from S;\\
            $\mathcal{N}_{i}$: the neighbouring concepts of $v_{i}$ in \textit{ConceptNet};\\
            \ForEach{$v_{j}$ in $\mathcal{N}_{i}$}{
                append $\left(v_{i}, r, v_{j}\right)$ into $T$;\\
            }
            select the top $K$ tail concepts which is most similar to the concepts in $V_{emo}$ from $T$;\\
            \ForEach{$\left(v_{i}, r, v_{j}\right)$ in $T$}{
                add $v_{j}$ and $r$ into $G$;\\
                \If{$v_{j}$ not in $V_{emo}$}{
                    push $v_{j}$ into $S$
                }
            }
        }
    }
\end{algorithm} 

\subsubsection{Graph-based Emotional Causality Encoder}\label{sec:graph-encoder}
Graph neural networks have been proved effective for producing graph-aware representations of nodes in the graph via integrating neighbours' information of each node \cite{zhou2018cnn}. Therefore, to obtain the representation of concepts and relations in the emotional causality graphs, we apply multi-layer GCN \cite{gcn} encoders to encode the emotional causality graphs. Moreover, following the idea of the TransE model \cite{transe}, we update a concept embedding with the subtraction between each neighbour concept embedding and the corresponding relation embedding to obtain the relation representation.

For each emotion clause, there is a constructed emotional causality graph $G=(V, R)$, where $V$ denotes concepts and $R$ denotes relations. The concepts in $V$ are initialized by pretrained word embeddings, and the relations in $R$ are initialized with randomly initialized relation-type embeddings. For each concept $v_{i}$, we update its embedding at the $(l+1)^{th}$ layer by aggregating its neighbours $\mathcal{N}_{i}$ including pairs of the concept and the relation liking to $v_{i}$:
\begin{equation}
h_{i}^{(l+1)}=\sigma\left(W_s^{(l)} h_{i}^{(l)}+\sum_{(j, r) \in N_{i}} \frac{1}{\left|\mathcal{N}_{i}\right|} W_n^{(l)} \left(h_j^{(l)}-h_r^{(l)}\right)\right),
\end{equation}
where $h_{i}^{(l)}$, $h_{j}^{(l)}$ and $h_{r}^{(l)}$ are the embeddings of node $v_i$, node $v_j$, and the relation between $v_i$ and $v_j$ at layer $(l)^{th}$; $W_s^{(l)}$ and $W_n^{(l)}$ are the two trainable parameter matrices specific to the layer $l^{t h}$; and $\sigma$ is a non-linear active function. The relation embedding is also updated at the $(l+1)^{t h}$ layer via a linear active function:
\begin{equation}
h_{r}^{(l+1)}=W_{R}^{(l)} h_{r}^{(l)}.
\end{equation}
After $L$ iterations, we are able to obtain $\left\{h_{v_{1}}^{(L)}, \ldots, h_{v_{|V|}}^{(L)}\right\}$, a set of concept representations, and $\left\{h_{r_{1}}^{(L)}, \ldots, h_{r_{|R|}}^{(L)}\right\}$, a set of relation representations.

\subsubsection{Transformer-based Conversation Context Encoder}\label{sec:trs-encoder}
We utilize a transformer encoder \cite{transformers,dehghani2018universal} for conversation context understanding. Same as previous works \cite{moel,majumder2020mime,li2019empgan}, we flatten all conversation context in $C$, and then add a special token $CLS$ at the beginning of the input, denoting the global memory of the whole sequence.
\begin{equation}
    \overline{C} =\left[CLS;C\right].
\end{equation}
After that, the input tokens are converted to word embeddings. To embody the influence of the emotional cause effect, we incorporate fixed causal embeddings into the input to show the importance of the clause in terms of their frequency as an emotion cause. In addition, we adopt dialogue state embeddings to distinguish situation text, the speaker's utterance and the listener's utterance. The representation of the encoder input is as follows:
\begin{equation}
    E^{C} = emb\left(\overline{C}\right) + emb_{cas}\left(\overline{C}\right) + emb_{utt}\left(\overline{C}\right).
\end{equation}
Here, $emb$, $emb_{cas}$, and $emb_{utt}$ represent the token embedding layer, causal embedding layer, and the dialogue state embedding layer, respectively. Finally, we use Transformer \cite{transformers} to encode the input.
\begin{equation}
    S^{C} = trs_{enc}\left(E^{C}\right),
\end{equation}
where ${S^{C}} \in \mathcal{R}^{{L_{c}\times d_{model}}}$, and $L_{C}$ is the length of the input sequence.

To better learn the context representation, we introduce an auxiliary task of emotion prediction, which is also adopted in previous methods \cite{moel,majumder2020mime,li2019empgan}. Specifically, we use the global memory of the whole sequence $S^{C}_{0}$ to predict the emotion label $q_{e}$ as follows:
\begin{gather}
    q = {S^{C}_{0}},\\
    q_{e}\sim P\left(e|C\right)=\sigma(W_eq). \label{eqn:mul}
\end{gather}

\subsection{Emotional Causality Augmented Decoder}\label{sec:decoder}
The emotional causality augmented decoder for empathetic response generation consists of two components, namely, a transformer-based decoder involving the overall emotional causality information and a graph-based decoder with multi-hop reasoning on the emotional causality graphs. These two decoders compute a generic vocabulary distribution and a graph-related concept distribution, respectively. And the final vocabulary distribution is a gated combination of these two vocabulary distributions. Under this setting, we can enhance the impact of emotional causality in both implicit and explicit ways.

\subsubsection{Transformer-based Decoder with Emotional Causality}\label{sec:trs-decoder}
To integrate emotional causality with the original transformer-based decoder, we first apply GRU \cite{gru} to obtain an emotional causality representation of the overall emotional causality information, and then introduce the emotional causality to the process of generating the generic vocabulary distribution.

For each conversation, there is a sequence of encoded emotional causality graphs. We obtain the representation of each emotional causality graph through average pooling of triple vectors in the graph $(v_{i}, r, v_{j}) \in \mathcal{G}_{n}$, which is
\begin{equation}
\mathscr{h}_{\mathcal{G}_{n}}=\sum_{\left(v_{i}, r, v_{j}\right) \in \mathcal{G}_{n}} \frac{1}{\left|\mathcal{G}_{n}\right|}\left[h_{i}^{(L)} ; h_{r}^{(L)} ; h_{j}^{(L)}\right],
\end{equation}
where $\left[;\right]$ represents the concatenation operation. With this sequence of emotional causality graph representations $\left\{\mathcal{G}_{1}, \ldots, \mathcal{G}_{N_{q}}\right\}$, we use a bidirectional \textit{Gate Recurrent Unit} (GRU) \cite{gru} to encode them. That is:
\begin{equation}
\begin{aligned}
    \overrightarrow{s_{n}}=G R U\left(\overrightarrow{s_{n-1}}, \mathscr{h}_{\mathcal{G}_{n}}\right),\\
    \overleftarrow{s_{n}}=G R U\left(\overleftarrow{s_{n-1}}, \mathscr{h}_{\mathcal{G}_{N_{q}-n+1}}\right).
\end{aligned}
\end{equation}
Then, we concatenate the forward last hidden state and the backward last hidden state of the GRU cells to combine the forward and backward representations of the emotional causality:
\begin{equation}
    s_{N_{q}}=\left[\overrightarrow{s_{N_{q}}} ; \overleftarrow{s_{N_{q}}}\right].
\end{equation}
Finally, we obtain the emotional causality vector, representing the overall emotional causality, through a linear transformation:
\begin{equation}
    \mathcal{H_{Q}}=\sigma(W_gs_{N_{q}}),
\end{equation}
where $W_g$ is the trainable parameter matrix.

Then we integrate the emotional causality vector to the transformer decoder. Similar to the context encoder, we use a special token $SOS$ together to represent the universal information of the decoder input $y_{<t}$. Then we append the emotional causality vector $\mathcal{H_{Q}}$ to this special vector. That is
\begin{gather}
    E_{t} = emb\left([y_{<t}]\right),\\
    s_{t}=trs_{dec}([emb(SOS)+\mathcal{H_{Q}};E_{t}], S^{C}). \label{eqn:decoder-state}
\end{gather}
Finally, we compute the generic vocabulary distribution:
\begin{equation}
    P\left(\mathcal{V} \mid \boldsymbol{s}_{t}, \mathcal{H_{Q}}\right)=softmax_{w\in \mathcal{V}}\left(W_{voc}s_{t}+b\right). \label{eqn:generic}
\end{equation}

\subsubsection{Graph-based Decoder with Multi-hop Reasoning}\label{sec:graph-decoder}
Inspired by the idea of \textit{multi-hop reasoning flow} \cite{ji2020language} and \textit{pointer-generator decoder} \cite{bauer2018commonsense}, we adopt a graph-based decoder, which computes the concept distribution with multi-hop reasoning on the emotional causality graphs. At each decoding step, we adjust the weight of each concept in the graph by combining its neighbouring evidences and the current decoder state. Initially, scores of all cause concepts are set to 1, while other concepts are assigned with a score of 0. Then, we broadcast information of scored concepts on the graph to update the unvisited concepts. For an unvisited concept $v_{i} \in V$ , $score(v)$ is computed by aggregating the evidences from its visited neighbouring concepts ${\mathcal{N}_{i}}^{in}$:
\begin{equation}
    score\left(v_{i}\right)=\sum\limits_{(v_{j},r)\in {\mathcal{N}_{i}}^{in}}\frac{1}{\left|{\mathcal{N}_{i}}^{in}\right|}\left(\gamma\cdot score(v_{j})+R\left(v_{j}, r, v_{i}\right)\right).
\end{equation}
Here, $\gamma$ is a discount factor and $R\left(v_{j}, r, v_{i}\right)$ is the triple relevance under the current decoder output, which can be computed as below:
\begin{gather}
    h_{\left(j,r,i\right)}=\left[h_{j}^{(L)} ; h_{r}^{(L)} ; h_{i}^{(L)}\right],\\
    R\left(v_{j},r,v_{i}\right)=\sigma\left(h_{\left(j,r,i\right)}W_{rel}s_{t}\right).
\end{gather}
After $L$-hop interactions, the distribution over the concepts is as follows:
\begin{equation}
    P\left(V \mid \boldsymbol{s}_{t}, \mathcal{G}\right)=softmax_{v\in V}\left(score\left( v \right) \right). \label{eqn:concept-dis}
\end{equation}

The final generation distribution conjoins the distribution over the concepts in the emotional causality graphs $V$ and the distribution over the standard vocabulary $\mathcal{V}$ with a soft gate $g_{t}$, which decides whether or not to refer to the emotional causality graphs:
\begin{gather}
    g_{t}=\sigma\left(W_{g}s_{t}\right),\\
    \begin{aligned} o_{t}\sim P\left(y_{t} \mid {y}_{<t}, C, \mathcal{G}, \mathcal{H_{Q}}\right) & = g_{t} \cdot P\left(V \mid \boldsymbol{s}_{t}, \mathcal{G}\right) \\ &+\left(1-g_{t}\right) \cdot P\left(\mathcal{V} \mid \boldsymbol{s}_{t}, \mathcal{H_{Q}}\right) \end{aligned}\label{vd}.
\end{gather}

\subsection{Training}\label{sec:loss}
To enhance the empathy of the model, we train the model via multi-task learning with an auxiliary task to predict the emotion label of the speaker. We compute the multi-task loss with the prediction results $q_{e}\sim P\left(e|C\right)$ which is compute in \equationautorefname~\eqref{eqn:mul}. 
\begin{equation}
    L^{e}=-\sum_{e=1}^{|E|}p_{e}log\left(q_{e}\right), \label{eqn:loss1}
\end{equation}
where $p_{e}$ is the real label probability of the speaker's emotion (1 for the true emotion label, and 0 for the others), and the $q_{e}$ is the predicted one (\cref{sec:trs-encoder}). 

The loss of response generation task is:
\begin{equation}\label{eqn:loss2}
    L^{g}=-\sum_{t=1}^{|Y|}\sum_{v=1}^{|\mathcal{V}|}{p_{t}}^{v}log\left({o_{t}}^{v}\right),
\end{equation}
where $Y$ is the target output, ${p_{t}}^{v}$ is the true probability of the token $w$ at the position $t$ in $Y$, and ${o_{t}}^{v}$ is the predicted one.

The total loss of the model is a weighted sum of these two components through a coefficient $\lambda$:
\begin{equation}\label{eqn:loss3}
    Loss=L^{g}+\lambda L^{e}.
\end{equation}

\section{Experiments}
\subsection{Dataset}
In this paper, we utilize \textsc{EmpatheticDialogues} \cite{empatheticdialogues}, a dataset of conversations grounded in specific emotional situations to conduct experiments. \tableautorefname~\ref{exm} shows an example from \textsc{EmpatheticDialogues}. This dataset consists of 25k crowdsourced one-on-one conversations between a speaker and a listener. Each dialogue is developed based on an emotion label and a given situation paragraph. There are 32 emotion categories and they are distributed in a balanced way. The speakers talk about their situations and the listeners attempt to understand the speakers' feelings and reply accordingly. At the training time, the emotional labels of the conversations are given in terms of the degree of empathy to facilitate multi-task learning, while the labels are not provided 
when testing the model and evaluating the generation performance.

\begin{table*}[width=.6\textwidth]\small
\centering
\caption{\textrm{An example from \textsc{EmpatheticDialogues}.}}\label{exm}
\begin{tabular}[width=.6\textwidth,cols=1,pos=ht]{p{.6\textwidth}}
\toprule
\textbf{Label:} Excited\\
\textbf{Situation:} My wife and I took a 3 week trip to New Zealand last year.\\
\textbf{Conversation:}\\
\textcolor{blue}{\textit{Speaker}}: My wife and I decided to go on an international trip last year, and it was my first trip out of the US. It was an amazing experience, and one of the best and most memorable things I've ever done.\\
\textcolor{red}{\textit{Listener:}} That's awesome! How helpful where the other people so that you could enjoy your trip? \\
\textcolor{blue}{\textit{Speaker:}} My wife was amazing. She booked the whole trip for us, and planned out some things for us to do while we were there. \\
\textcolor{red}{\textit{Listener:}} Must have been super fun. \\
\bottomrule
\end{tabular}
\end{table*}

\subsection{Comparison Models}
We compare our \textit{GREC} with the following models for empathetic response generation:\\
\textbf{Transformer with Multitask (Multi-TRS)}. This is a universal transformer encoder-decoder structure model \cite{transformers}. We train this model in a multi-task learning setting, where the loss is calculated by \equationautorefname~\eqref{eqn:loss1}--\eqref{eqn:loss3}.\\ 
\textbf{MoEL}. The mixture of Empathetic Listeners (MoEL) \cite{moel} is a transformer-based model consisting of multiple emotion-specific decoders and a meta decoder. Each emotion-specific decoder is optimized to react to a certain emotion, while the meta decoder integrates evidences from different decoders and generates the final empathetic reply. We implement it using the same parameters and the values reported in \cite{moel}.  \\
\textbf{MIME}. MIME \cite{majumder2020mime} designs an empathetic response generation model with the consideration of polarity-based emotion clusters and emotional mimicry. It attempts to achieve an appropriate balance of emotions in positive and negative emotion groups. Based on MoEL \cite{moel}, MIME introduces stochasticity into the emotion mixture for various empathetic responses. \\
\textbf{EmpDG}. This model \cite{li2019empgan} exploits both dialogue-level and token-level emotion to capture the nuances of the speaker's emotion. It also involves the speaker's feedback to achieve better expression of empathy with an interactive adversarial learning framework.

\subsection{Experiment Setup}
We use \textit{Pytorch}\footnote{https://pytorch.org/} to implement the proposed model. The word embedding is initialized with pretrained Glove vectors\footnote{https://github.com/stanfordnlp/GloVe} of dimension 300 \cite{pennington2014glove} and shared by the transformer encoder, GCN encoder, and the projection layer of the decoder. For the transformer encoder and decoder, the layer number is set to 6, and the number of attention heads is set to 8. The embedding dimensions of the query, key, and value are set to 40. We replace the position wise feedforward sub-layer with a 1D convolution with 50 filters of width 3. The causal embeddings to denote the importance of the clauses in the conversation context are randomly initialized and fixed during training, and the embedding dimension is set to 300 empirically. For RTHN, which is used to extract the emotional causal relationships from the conversation, we follow the original paper \cite{rthn} and pretrain an emotion causality detection model based on an annotated relational emotion dataset, namely \textsc{REMAN}\footnote{https://webanno.github.io/webanno/use-case-gallery/reman/} \cite{Kim2018}. We finally use a \textit{RTHN} model that has F1 of 0.9481. We use Adam for optimization \cite{kingma2014adam}. The batch size is 32 and the initial learning rate is 0.1. The learning rate is updated every 500 steps with a decay rate of 0.1 until the learning rate is 1e-5. The beam size is set to 5. For statistical significance tests, we run the model 10 times.

For a fair comparison, the parameter settings and training processes of all state-of-the-art models are the same as suggested in their original papers and implementations. 

\subsection{Evaluation Metrics}
We evaluate our models from three aspects, i.e., automatic evaluation, human ratings, and human A/B test.

\textbf{Automatic Evaluation}. We use per-word perplexity (PPL) \cite{bahl1983maximum} and BLEU \cite{papineni2002bleu} to measure the performance of response generation. PPL measures the confidence to generate the responses. BLEU evaluates the coherency of the generated responses with golden responses \cite{liu2016not}. Lower PPL and higher BLEU are expected. We use \textit{multi-bleu.perl}\footnote{https://github.com/google/seq2seq/} \cite{Britz:2017} to compute the BLEU scores. 

\textbf{Human Ratings}. For qualitative evaluation, we use manual ratings to compare \textit{GREC} and other models. We first randomly sample 120 instances from the test dataset. Then, we score these responses from the following aspects: \textbf{empathy} (whether the response shows an understanding of the feelings of the speaker), \textbf{relevance} (whether the response is relevant to the conversation) and \textbf{fluency} (whether the response is fluent and natural). The scores range from 1 to 5, where 1 is the lowest and 5 is the highest. We evaluate each metric independently.

\textbf{Human A/B Test}. In order to compare the overall performances of different models, we adopt human A/B task. Given two models A and B -- \textit{GREC vs. {Multi-TRS, MoEL, MIME}} in our case, we ask the annotator to pick the model with the better responses. The annotator can select a \textit{Tie} if the responses are both good or both bad. 

\section{Results and Analysis}
\subsection{Experimental Results}
The results of the automatic evaluation and human ratings are shown in \tabref{results}. We find that our model outperforms other models in most of the evaluation metrics. Our model achieves the lowest PPL and the highest BLEU among all models, which indicates that responses generated by our model are more human-like. For the human rating results, \textbf{fluency} of different models are marginal. However, it is obvious that \textit{GREC} obtains the highest scores in terms of \textbf{empathy} and \textbf{relevance}. We think this is achieved because of the incorporation of emotional causality. Although the models MoEl, MIME and EmpDG also employ emotion wisely, they merely utilize emotion labels or emotional words to represent the emotion of the speaker. However, emotional causality helps the model to understand how the emotion of the speaker emerges. Emotional causality provides rich semantic relations between the cause and the emotion, which facilities the understanding of the speaker's emotion. The multi-hop reasoning also emphasizes relevant information in the emotional causality graphs. Therefore responses generated by \textit{GREC} are more empathetic responses and relevant.

\begin{table}[width=1.0\linewidth]\footnotesize
\caption{Results of automatic evaluation and manual evaluation. Asterisk symbols represent statistical significance according to student t-test the best comparison model, with $p<0.05(*)$ and $p<0.01(**)$, respectively.}\label{results}
\begin{tabular}[width=\linewidth,cols=6,pos=ht]{>{\raggedright\arraybackslash}p{.19\linewidth}>{\centering\arraybackslash}p{.08\linewidth}>{\centering\arraybackslash}p{.12\linewidth}>{\raggedright\arraybackslash}p{.09\linewidth}>{\raggedright\arraybackslash}p{.11\linewidth}>{\raggedright\arraybackslash}p{.08\linewidth}}
\toprule
{Methods} &\multicolumn{2}{L}{Automatic evaluation}&
\multicolumn{3}{C}{Manual evaluation}\\
\cmidrule(lr){2-3} \cmidrule(lr){4-6}
& PPL$\downarrow$ & BLEU$\uparrow$ & Empathy$\uparrow$ & Relevance$\uparrow$ & Fluency$\uparrow$\\
\hline
    \textit{Multi-TRS}&33.00&2.95&2.90&2.94&3.70\\
    \textit{MoEL}&35.35&2.64&2.91&2.91&3.77\\
    \textit{MIME}&33.05&2.72&3.06&3.13&\textbf{3.79}\\
    \textit{EmpDG}&56.00&1.70&3.05&2.87&3.68 \\
    \textit{GREC (Ours)}&\textbf{32.66}&\textbf{3.16**}&\textbf{3.21*}&\textbf{3.32*}&3.78\\
\bottomrule
\end{tabular}
\end{table}

From \tabref{ABtest}, responses generated by \textit{GREC} are more often preferred by human judges in general than the responses from \textit{Multi-TRS}, \textit{MoEL}, and \textit{MIME}. This also indicates that our model is able to generate more empathetic and more relevant responses. 

\begin{table}[width=\linewidth]\footnotesize
\caption{Human A/B test results for \textit{GREC vs. {Multi-TRS, MoEL and MIME}}}\label{ABtest}
\begin{tabular}[width=\linewidth,cols=4,pos=ht]{>{\raggedright\arraybackslash}p{.2\linewidth}>{\raggedright\arraybackslash}p{.2\linewidth}>{\raggedright\arraybackslash}p{.2\linewidth}>{\raggedright\arraybackslash}p{.2\linewidth}}
\toprule
\textit{GREC vs.}& \textit{GREC} wins& \textit{GREC} loses&Tie\\
\midrule
     \textit{Multi-TRS}&50.0\%&22.5\%&27.5\%\\
     \textit{MoEL}&48.3\%&20.8\%&30.8\%\\
     \textit{MIME}&35.0\%&27.5\%&37.5\%\\
     \textit{EmpDG}&62.7\%&11.7\%& 25.6\%\\
\bottomrule
\end{tabular}
\end{table}

\subsection{Analysis and Discussion}
\subsubsection{Ablation Study} \label{as}
To verify the effect and the contribution of each component of \textit{GREC}, we conduct the ablation study. We consider the following three settings.

(1) \textbf{w/o graph}. We remove the graph structure of emotional causality, and involve emotional causality in the format of text. Hence, we delete the emotional causality construction (\sectionautorefname~\cref{sec:construct}). For emotional causality understanding, we use a sequence of emotional cause clause contexts extracted from the conversation context, and apply RNNs to encode them instead of the GCN encoder in \sectionautorefname~\cref{sec:graph-encoder}. We retain the implicit and explicit use of emotional causality when decoding. Instead of the concept distribution computation in \equationautorefname~\eqref{eqn:concept-dis}, we compute the weight of the emotion concepts and the cause concepts $\mathcal{V}$ by
\begin{equation}
     P\left(V \mid \boldsymbol{s}_{t},\mathcal{V}\right)=softmax_{w\in \mathcal{V}}\left(emb(w)s_{t}+b\right).
\end{equation}

(2) \textbf{w/o implicit decoding}. We ablate the implicit use of the emotional causality by deleting the involvement of the emotional causality in the transformer decoder in \sectionautorefname~\cref{sec:trs-decoder} (i.e., \textit{w/o encoder}). We replace \equationautorefname~\eqref{eqn:decoder-state} by
\begin{equation}
    s_{t}=trs_{dec}([emb(SOS);E_{t}], S^{C}).
\end{equation}

(3) \textbf{w/o explicit reasoning}. We ablate the explicit use of the emotional causality by removing the graph-based decoder with multi-hop reasoning in \sectionautorefname~\cref{sec:graph-decoder}.  The final vocabulary distribution is calculated by \equationautorefname~\eqref{eqn:generic} only.

\begin{table}[width=\linewidth]\scriptsize
\caption{Results of automatic evaluation and manual evaluation for \textit{Hierarchical Emotional Causal RNN}.}\label{ab}

\begin{tabular}[width=\linewidth,cols=6,pos=ht]{>{\raggedright\arraybackslash}p{.26\linewidth}>{\centering\arraybackslash}p{.08\linewidth}>{\centering\arraybackslash}p{.08\linewidth}>{\raggedright\arraybackslash}p{.08\linewidth}>{\raggedright\arraybackslash}p{.1\linewidth}>{\raggedright\arraybackslash}p{.08\linewidth}}
\toprule
{Methods} &\multicolumn{2}{L}{Automatic evaluation}&
\multicolumn{3}{C}{Manual evaluation}\\
\cmidrule(lr){2-3} \cmidrule(lr){4-6}
& PPL$\downarrow$ & BLEU$\uparrow$ & Empathy$\uparrow$ & Relevance$\uparrow$ & Fluency$\uparrow$\\
\midrule
\textit{GREC}&32.66&3.16&3.21&3.32&3.78\\
\midrule
\textit{w/o graph}&33.93&2.20&2.92&2.94&3.75\\
\textit{w/o implicit decoding}&33.50&3.05&3.13&3.17&3.67\\
\textit{w/o explicit reasoning}&32.94&2.81&3.02&3.02&3.85\\
\bottomrule
\end{tabular}
\end{table}

The results in \tableautorefname~\ref{ab} show that all components contribute to \textit{GREC} because removing any of them decreases the overall performance. Moreover, each component contributes to \textit{GREC} from different aspects. Based on the observation and the design of our model, we have the following analysis. 

(1). Emotional causality graphs contribute most to our model. The performance drops in both \textbf{empathy} and \textbf{relevance} significantly by $9.03\%$ and $11.45\%$ respectively in the absence of the graph structure. Graph structures with external commonsense knowledge can provide a better representation of the emotional causality. Although the leverage of emotional cause clause can also provide evidence for emotion understanding, which outperforms Multi-TRS and MoEL in the aspect of \textbf{empathy} and \textbf{Relevance}, the semantic relationship between the cause and emotion is more abundant in the emotional causality graphs. Besides, emotional causality graphs involve more information for the decoder with multi-hop reasoning. 

(2). Decoding with the emotional causality in an implicit way also contributes to our model. Compared to the other two ablation models, the performance is decreased less in most of the evaluation metrics, however, the \textbf{fluency} decreases significantly. On the one hand, the transformer decoder with emotional causality is an implicit way to use emotional causality, therefore it cannot change the performance remarkably. On the other hand, it provides a chance where the emotional causality can influence the computation of generic vocabulary distribution, making the involvement of emotional causality more natural.

(3). Graph-based decoding with multi-hop reasoning also has a great contribution to \textbf{GREC}. This component reasons out important information from the emotional causality graphs, which increases empathetic and relevant expression. However, the graph-based decoder with multi-hop reasoning has a negative influence on \textbf{fluency}. One reasonable explanation is that emotional causality graphs contain not only useful information, but also noises, which decrease the \textbf{fluency} of the responses. In future work, it is a meaningful direction to attend more on the useful evidences, and reduce the influence of noises in the graphs.

\subsubsection{Case Study}
\begin{figure*}
	\centering
		\includegraphics[width=\textwidth]{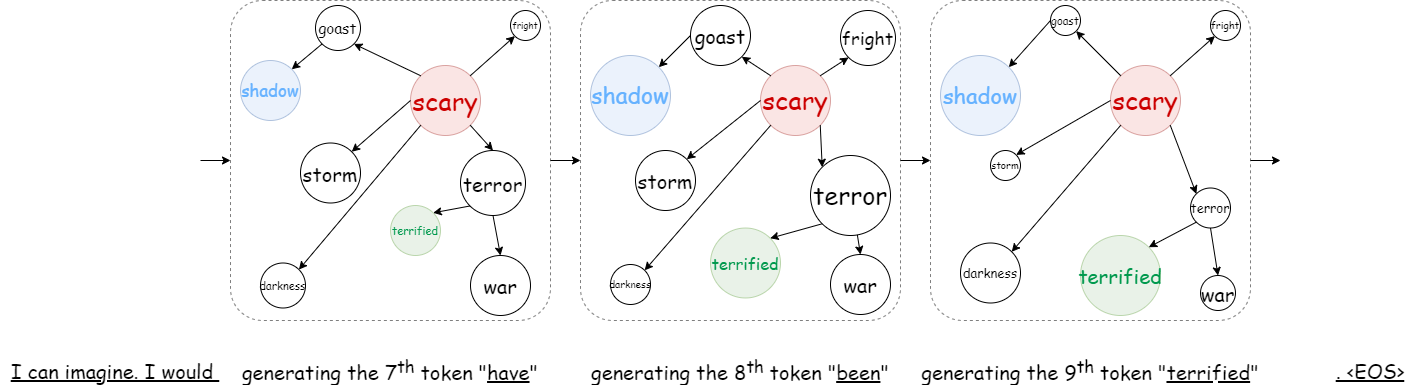}
	\caption{This is a piece from the emotional causality graph for \textit{``It was so scary \dots an intruder!"}, which shows the attention weights of the concepts vary over the decoding steps. \textsc{\textcolor{red}{\textit{Red circles}}} represent the cause concept, \textsc{\textcolor{CornflowerBlue}{\textit{blue circles}}} represent the emotion concept, and \textsc{\textcolor{OliveGreen}{\textit{green circles}}} represent the concept showing in the final generated response. The size of the circle corresponds to the attention weight for the concept in the current decoding step.}
	\label{graph}
\end{figure*}

We provide some cases to show the effectiveness of \textit{GREC}. \ref{tab:case} shows two instances from the test data. From the results, \textit{GREC} can capture the emotion of the speaker more accurately, and respond in a more natural way, compared with other models. In both these two conversations, \textit{GREC} generates responses with appropriate emotion and expression. For example, it can use the verb ``get it back" and ``complain" as the response to a disgusting attitude. Such expression is pertinent and specific in this conversation, showing a deeper understanding of the speaker's emotion. However, other models prefer to using emotional words to express the emotion, which is not natural and can be used in a wide range of scenarios. It can also capture the concepts that bring the emotion of the speaker. In the second conversation, \textit{GREC} talks about ``shadow", which is the core content of this conversation. Besides, \textit{GREC} usually performs better in or after the second turn of a conversation compared to other models, which is also reflected in these two cases. We believe that the emotional causality graphs carry important information in the conversation context, and allow such information to have an impact on the current turn. 
\begin{table*}[width=.86\textwidth]\footnotesize
\caption{Responses generated by \textit{GREC}. These utterances are from continuous dialogues. The gold responses are also displayed. Terms showing empathy are labelled in \textcolor{OliveGreen}{\textit{green}}.}\label{tab:case}
\centering
\begin{tabular}[width=.6\textwidth,cols=2,pos=ht]{>{\raggedright\arraybackslash}p{.1\textwidth} >{\raggedright\arraybackslash}p{.5\textwidth}}
\toprule
\multicolumn{2}{>{\raggedright\arraybackslash}p{.85\textwidth}}{Conversation 1}\cr
\midrule
\multicolumn{2}{>{\raggedright\arraybackslash}p{.85\textwidth}}{Label: DISGUSTED}\\
\midrule
\multicolumn{2}{>{\raggedright\arraybackslash}p{.85\textwidth}}{Situation: I once went to a pizza place and found a hair on my pizza. I was grossed out.}\\
\midrule
\textcolor{blue}{\textbf{Speaker}}&I had pizza at a restaurant a couple days ago and found a hair on my pizza!\\
\textcolor{red}{\textbf{Gold}}&It is \textcolor{OliveGreen}{\textit{disgusting}}. What did you do?\\
\textbf{Transformer}&What did you do ?\\
\textbf{GREC}&That is so \textcolor{OliveGreen}{\textit{gross}}! Did you get it back?\\
\textcolor{blue}{\textbf{Speaker}}&I told the waiter and demanded a refund! It was gross.\\
\textcolor{red}{\textbf{Gold}}&You did good. You should have called the manager to tell him know.\\
\textbf{Transformer}&Wow, that is \textcolor{OliveGreen}{\textit{gross}}. I hope you were able to fix it .\\
\textbf{GREC}&It is so \textcolor{OliveGreen}{\textit{gross}}. Did you \textcolor{OliveGreen}{\textit{complain}} to the manager ?\\
\midrule
\multicolumn{2}{>{\raggedright\arraybackslash}p{.85\textwidth}}{Conversation 2}\cr
\midrule
\multicolumn{2}{>{\raggedright\arraybackslash}p{.85\textwidth}}{Label: TERRIFIED}\\
\midrule
\multicolumn{2}{>{\raggedright\arraybackslash}p{.85\textwidth}}{\makecell[l]{Situation: One night my children and I came home, and I saw a shadow inside the home when I came inside. \\My heart jumped into my throat and you could hear my scream in the next town for fear!}}\\
\midrule
\textcolor{blue}{\textbf{Speaker}}&One night when my children and I came home, I saw a shadow inside! I was so fearful and afraid!\\
\textcolor{red}{\textbf{Gold}}&Oh that can be \textcolor{OliveGreen}{\textit{scary}}. I don't watch horror movies because I know that I'll do just that!\\
\textbf{Transformer}&\textcolor{OliveGreen}{\textit{Terrified}}, oh no! I am so sorry to hear that.\\
\textbf{GREC}&Oh no! I \textcolor{OliveGreen}{\textit{hate}} when that happens to me. What was it?\\
\textcolor{blue}{\textbf{Speaker}}&It was so scary! Luckily, it was actually my shadow! The light made my shadow show up, but it first appeared to be an intruder!\\
\textcolor{red}{\textbf{Gold}}&Well I'm glad you were ok! Sometimes it's not just a shadow.\\
\textbf{Transformer}&\textcolor{OliveGreen}{\textit{Terrified}}, I am sure it was a \textcolor{OliveGreen}{\textit{scary movie}}.\\
\textbf{GREC}&I can imagine. I would have been \textcolor{OliveGreen}{\textit{terrified}}. \\
\bottomrule
\end{tabular}
\end{table*}
To figure out how the information in the emotional causality helps the empathetic response generation, we display the scores of the concepts in the emotional causality graph when generating a sentence. \ref{graph} shows an emotional causality graph for the second utterance of the speaker. The related concepts, which are suitable for this conversation, are inferred and selected. Moreover, the attention weights of the concepts varying over the decoding steps are reflected in this figure. The attention weights for the trivial terms, e.g.,~``war" and ``storm", decrease, while weights for necessary terms, such as ``terrified" and ``darkness", increase over steps. In addition, \textit{``terrified"} is even selected in the response. This indicates that the model can reason out important information from the emotional causality graphs for empathetic response generation through a multi-hop strategy. Besides, the concepts with higher scores are deduced with flows from the cause concepts. This indeed enhances model transparency and explainability.

\section{Conclusions}
In this work, we incorporate the emotional causality into empathetic response generation. To implement this idea, we propose a novel graph-based multi-hop reasoning model, called \textit{GREC}. Instead of simple employment of the superficial emotional information, such as emotion labels and emotional words, \textit{GREC} explores how the emotion emerges and provides a deeper understanding of the speaker's emotion. With a typical transformer Seq2Seq model as the skeleton, we augment the encoder and decoder with emotional causality. For emotional causality representation and understanding, we construct emotional causality graphs through multi-hop reasoning over ConceptNet, and apply a multi-layer GCN to encode them. For empathetic response generation, we involve the emotional causality information into the transformer decoder and design a graph-based decoder with multi-hop reasoning on the emotional causality graphs. We conduct experiments on a crowdsourced dataset, i.e., \textsc{EmpatheticDialogues}, and empirically prove that \textit{GREC} outperforms competitive models including those state-of-the-art. We show that graph-based reasoning provides better model transparency and explainability in the ablation study and the case study.

\section*{Acknowledgement}
The work described in this paper was supported by Research Grants Council of Hong Kong (PolyU 152040/18E, PolyU 15207920), National Natural Science Foundation of China (62076212) and PolyU (ZVQ0).

\bibliographystyle{ieeetr}
\bibliography{refs}

\begin{thebibliography}{10}

\bibitem{eide2004listening}
H.~Eide, R.~Frankel, A.~C.~B. Haaversen, K.~A. Vaupel, P.~K. Graugaard, and
  A.~Finset, ``Listening for feelings: identifying and coding empathic and
  potential empathic opportunities in medical dialogues,'' {\em Patient Educ.
  Couns.}, vol.~54, no.~3, pp.~291--297, 2004.

\bibitem{perez2017understanding}
V.~P\'erez-Rosas, R.~Mihalcea, K.~Resnicow, S.~Singh, and L.~An, {\em
  Understanding and Predicting Empathic Behavior in Counseling Therapy},
  pp.~1426--1435.
\newblock 2017.

\bibitem{zhou2020design}
L.~Zhou, J.~Gao, D.~Li, and H.-Y. Shum, ``The design and implementation of
  xiaoice, an empathetic social chatbot,'' {\em Computational Linguistics},
  vol.~46, no.~1, pp.~53--93, 2020.

\bibitem{moel}
Z.~Lin, A.~Madotto, J.~Shin, P.~Xu, and P.~Fung, {\em MoEL: Mixture of
  Empathetic Listeners}, pp.~121--132.
\newblock 2019.

\bibitem{majumder2020mime}
N.~Majumder, P.~Hong, S.~Peng, J.~Lu, D.~Ghosal, A.~Gelbukh, R.~Mihalcea, and
  S.~Poria, {\em MIME: MIMicking Emotions for Empathetic Response Generation},
  pp.~8968--8979.
\newblock 2020.

\bibitem{li2019empgan}
Q.~Li, H.~Chen, Z.~Ren, P.~Ren, Z.~Tu, and Z.~Chen, {\em EmpDG:
  Multi-resolution Interactive Empathetic Dialogue Generation}, pp.~4454--4466.
\newblock 2020.

\bibitem{wierzbicka1999emotions}
A.~Wierzbicka.
\newblock 1999.

\bibitem{whatisemo}
W.~James, ``{{WHAT} {IS} AN {EMOTION}?},'' {\em Mind}, vol.~os-IX, no.~34,
  pp.~188--205, 1884.

\bibitem{gcn}
T.~N. Kipf and M.~Welling, {\em Semi-Supervised Classification with Graph
  Convolutional Networks}.
\newblock 2017.

\bibitem{ma2020survey}
Y.~Ma, K.~L. Nguyen, F.~Z. Xing, and E.~Cambria, ``A survey on empathetic
  dialogue systems,'' {\em Information Fusion}, vol.~64, pp.~50--70, 2020.

\bibitem{shi2018sentiment}
W.~Shi and Z.~Yu, {\em Sentiment Adaptive End-to-End Dialog Systems},
  pp.~1509--1519.
\newblock 2018.

\bibitem{zhong2019affect}
P.~Zhong, D.~Wang, and C.~Miao, 2019.

\bibitem{ecm}
H.~Zhou, M.~Huang, T.~Zhang, X.~Zhu, and B.~Liu, {\em Emotional Chatting
  Machine: Emotional Conversation Generation with Internal and External
  Memory}, pp.~730--739.
\newblock 2018.

\bibitem{song2019generating}
Z.~Song, X.~Zheng, L.~Liu, M.~Xu, and X.~Huang, {\em Generating Responses with
  a Specific Emotion in Dialog}, pp.~3685--3695.
\newblock 2019.

\bibitem{mojitalk}
X.~Zhou and W.~Y. Wang, {\em MojiTalk: Generating Emotional Responses at
  Scale}, pp.~1128--1137.
\newblock 2018.

\bibitem{empatheticdialogues}
H.~Rashkin, E.~M. Smith, M.~Li, and Y.-L. Boureau, {\em Towards Empathetic
  Open-domain Conversation Models: A New Benchmark and Dataset},
  pp.~5370--5381.
\newblock 2019.

\bibitem{lin2019caire}
Z.~Lin, P.~Xu, G.~I. Winata, F.~B. Siddique, Z.~Liu, J.~Shin, and P.~Fung,
  2020.

\bibitem{torres2019spoken}
M.~I. Torres, J.~M. Olaso, N.~Glackin, R.~Justo, and G.~Chollet, {\em A spoken
  dialogue system for the empathic virtual coach}, pp.~259--265.
\newblock 2019.

\bibitem{fung2018empathetic}
P.~Fung, D.~Bertero, P.~Xu, J.~H. Park, C.-S. Wu, and A.~Madotto, 2018.

\bibitem{alm2005emotions}
C.~O. Alm, D.~Roth, and R.~Sproat, {\em Emotions from Text: Machine Learning
  for Text-based Emotion Prediction}, pp.~579--586.
\newblock 2005.

\bibitem{aman2007identifying}
S.~Aman and S.~Szpakowicz, {\em Identifying expressions of emotion in text},
  pp.~196--205.
\newblock 2007.

\bibitem{bakshi2016opinion}
R.~K. Bakshi, N.~Kaur, R.~Kaur, and G.~Kaur, {\em Opinion mining and sentiment
  analysis}, pp.~452--455.
\newblock 2016.

\bibitem{lee2010text}
S.~Y.~M. Lee, Y.~Chen, and C.-R. Huang, {\em A Text-driven Rule-based System
  for Emotion Cause Detection}, pp.~45--53.
\newblock 2010.

\bibitem{lee2010emotion}
S.~Y.~M. Lee, Y.~Chen, S.~Li, and C.-R. Huang, {\em Emotion Cause Events:
  Corpus Construction and Analysis}.
\newblock 2010.

\bibitem{gui-etal-2016-event}
L.~Gui, D.~Wu, R.~Xu, Q.~Lu, and Y.~Zhou, {\em Event-Driven Emotion Cause
  Extraction with Corpus Construction}, pp.~1639--1649.
\newblock 2016.

\bibitem{xia2019emotion}
R.~Xia and Z.~Ding, {\em Emotion-Cause Pair Extraction: A New Task to Emotion
  Analysis in Texts}, pp.~1003--1012.
\newblock 2019.

\bibitem{gui2017question}
L.~Gui, J.~Hu, Y.~He, R.~Xu, Q.~Lu, and J.~Du, {\em A question answering
  approach to emotion cause extraction}, pp.~1593--1602.
\newblock 2017.

\bibitem{rthn}
R.~Xia, M.~Zhang, and Z.~Ding, {\em RTHN: A RNN-Transformer Hierarchical
  Network for Emotion Cause Extraction}, pp.~5285--5291.
\newblock 2019.

\bibitem{ji2020survey}
S.~Ji, S.~Pan, E.~Cambria, P.~Marttinen, and P.~S. Yu, ``A survey on knowledge
  graphs: Representation, acquisition, and applications,'' {\em IEEE
  Transactions on Neural Networks and Learning Systems}, pp.~1--21, 2021.

\bibitem{zhou2018commonsense}
H.~Zhou, T.~Young, M.~Huang, H.~Zhao, J.~Xu, and X.~Zhu, {\em Commonsense
  Knowledge Aware Conversation Generation with Graph Attention},
  pp.~4623--4629.
\newblock 2018.

\bibitem{lin2019kagnet}
B.~Y. Lin, X.~Chen, J.~Chen, and X.~Ren, {\em KagNet: Knowledge-Aware Graph
  Networks for Commonsense Reasoning}, pp.~2829--2839.
\newblock 2019.

\bibitem{bauer2018commonsense}
L.~Bauer, Y.~Wang, and M.~Bansal, {\em Commonsense for Generative Multi-Hop
  Question Answering Tasks}, pp.~4220--4230.
\newblock 2018.

\bibitem{liu-etal-2019-knowledge}
Z.~Liu, Z.-Y. Niu, H.~Wu, and H.~Wang, {\em Knowledge Aware Conversation
  Generation with Explainable Reasoning over Augmented Graphs}, pp.~1782--1792.
\newblock 2019.

\bibitem{ji2020language}
H.~Ji, P.~Ke, S.~Huang, F.~Wei, X.~Zhu, and M.~Huang, {\em Language Generation
  with Multi-Hop Reasoning on Commonsense Knowledge Graph}, pp.~725--736.
\newblock 2020.

\bibitem{zhao2020graph}
L.~Zhao, J.~Xu, J.~Lin, Y.~Zhang, H.~Yang, and X.~Sun, ``Graph-based multi-hop
  reasoning for long text generation,'' 2020.

\bibitem{zhou2018cnn}
J.~Zhou, G.~Cui, S.~Hu, Z.~Zhang, C.~Yang, Z.~Liu, L.~Wang, C.~Li, and M.~Sun,
  ``Graph neural networks: A review of methods and applications,'' {\em AI
  Open}, vol.~1, pp.~57--81, 2020.

\bibitem{transe}
A.~Bordes, N.~Usunier, A.~Garc{\'\i}a{-}Dur\'an, J.~Weston, and O.~Yakhnenko,
  2013.

\bibitem{transformers}
A.~Vaswani, N.~Shazeer, N.~Parmar, J.~Uszkoreit, L.~Jones, A.~N. Gomez,
  L.~Kaiser, and I.~Polosukhin, 2017.

\bibitem{dehghani2018universal}
M.~Dehghani, S.~Gouws, O.~Vinyals, J.~Uszkoreit, and L.~Kaiser, {\em Universal
  Transformers}.
\newblock 2019.

\bibitem{gru}
K.~Cho, B.~{van Merri\"enboer}, C.~Gulcehre, D.~Bahdanau, F.~Bougares,
  H.~Schwenk, and Y.~Bengio, {\em Learning Phrase Representations using RNN
  Encoder{--}Decoder for Statistical Machine Translation}, pp.~1724--1734.
\newblock 2014.

\bibitem{pennington2014glove}
J.~Pennington, R.~Socher, and C.~Manning, {\em GloVe: Global Vectors for Word
  Representation}, pp.~1532--1543.
\newblock 2014.

\bibitem{Kim2018}
E.~Kim and R.~Klinger, {\em Who Feels What and Why? Annotation of a Literature
  Corpus with Semantic Roles of Emotions}, pp.~1345--1359.
\newblock 2018.

\bibitem{kingma2014adam}
D.~P. Kingma and J.~Ba, 2015.

\bibitem{bahl1983maximum}
L.~R. Bahl, F.~Jelinek, and R.~L. Mercer, ``A maximum likelihood approach to
  continuous speech recognition,'' {\em IEEE transactions on pattern analysis
  and machine intelligence}, no.~2, pp.~179--190, 1983.

\bibitem{papineni2002bleu}
K.~Papineni, S.~Roukos, T.~Ward, and W.-J. Zhu, {\em Bleu: a Method for
  Automatic Evaluation of Machine Translation}, pp.~311--318.
\newblock 2002.

\bibitem{liu2016not}
C.-W. Liu, R.~Lowe, I.~Serban, M.~Noseworthy, L.~Charlin, and J.~Pineau, {\em
  How NOT To Evaluate Your Dialogue System: An Empirical Study of Unsupervised
  Evaluation Metrics for Dialogue Response Generation}, pp.~2122--2132.
\newblock 2016.

\bibitem{Britz:2017}
D.~Britz, A.~Goldie, M.-T. Luong, and Q.~Le, {\em Massive Exploration of Neural
  Machine Translation Architectures}, pp.~1442--1451.
\newblock 2017.

\end{thebibliography}

\end{document}